\pdfoutput=1

\documentclass[11pt]{article}

\usepackage[final]{acl}

\usepackage{times}
\usepackage{latexsym}

\usepackage[T1]{fontenc}
\usepackage[utf8]{inputenc}
\usepackage[T2A,LAE,T1]{fontenc}
\usepackage[arabic,USenglish]{babel}
\usepackage{subcaption}

\usepackage{microtype}

\usepackage{inconsolata}

\usepackage{graphicx}

\usepackage{todonotes}
\usepackage{booktabs}
\usepackage{ragged2e}
\usepackage{array}
\usepackage{multirow}
\usepackage{amsmath}
\usepackage[shortlabels]{enumitem}

\newcommand{\ARA}[1]{\begin{scriptsize}\AR{#1}\end{scriptsize}}

\title{LLM Alignment for the Arabs: A Homogenous Culture or Diverse Ones?}

\author{Amr Keleg \\
Institute for Language, Cognition and Computation \\
School of Informatics, University of Edinburgh \\
  \texttt{a.keleg@sms.ed.ac.uk}}

\begin{document}
\maketitle
\begin{abstract}
Large language models (LLMs) have the potential of being useful tools that can automate tasks and assist humans. However, these models are more fluent in English and more aligned with Western cultures, norms, and values. Arabic-specific LLMs are being developed to better capture the nuances of the Arabic language, as well as the views of the Arabs. Yet, Arabs are sometimes assumed to share the same culture. In this position paper, I discuss the limitations of this assumption and provide preliminary thoughts for how to build systems that can better represent the cultural diversity within the Arab world. The invalidity of the cultural homogeneity assumption might seem obvious, yet, it is widely-adopted in developing multilingual and Arabic-specific LLMs. I hope that this paper will encourage the NLP community to be considerate of the cultural diversity within various communities speaking the same language.
\end{abstract}

\section{Introduction}
Even in the global world we live in, people residing in different parts of the world nourish different ideas, have different interests, and face different challenges. These differences can be too extreme to the extent that people could be considered to be living in totally distinct worlds (\citealp[p. 209]{2598291a-d0b6-399c-a4fa-1f25e75dadc3} as cited in \citealp[p. 3]{bird-2024-must}).
For instance, \citet{kirk2024prism} found that US participants questioned Large Language Models (LLMs) about abortion more than non-US ones. People from different regions can also have different perceptions of the same topic, as exemplified by English speakers from the US, UK, Singapore, Kenya, and South Africa disagreeing on what counts as Hate Speech \cite{lee-etal-2024-exploring-cross}. All these differences could be attributed to the cultural diversity among various communities across the world.

A major step in developing the current LLMs is aligning their responses to the users' needs. With the popularized one-model-fits-all paradigm, it is challenging to build models that can produce personalized responses that appeal to people of different demographics \cite{kirk2024prism}.
Current models tend to generate responses that better match the expectations of Western users \cite{cao-etal-2023-assessing, naous-etal-2024-beer, wang-etal-2024-countries, alkhamissi-etal-2024-investigating, ryan-etal-2024-unintended, mihalcea2024aiweirdwayai}.
Moreover, the views of Arabs---one group of many underrepresented non-Western communities---tend to be ignored,\footnote{Despite the attempt of curating model alignment data from different multi-cultural demographics, the PRISM Alignment dataset \cite{kirk2024prism} had only 51 participants (out of 1,500) who reported that they reside in the Middle East, out of which 47 reside in Israel, 2 in Turkey, and 1 in each of Sudan and Kuwait. Moreover, only 14 participants self-reported themselves as \textit{Middle Eastern/ Arab}.} sometimes unconsciously and other times with deliberate intent, putting people of these communities at a higher risk of discrimination \cite{alimardani2021digital,10.1145/3544548.3581538,walid_chi_2025}.

Arabic is privileged by having (a) a community of Arab NLP experts \cite{ws-2014-emnlp,ws-2015-arabic,ws-2017-arabic,ws-2019-arabic,wanlp-2020-arabic,wanlp-2021-arabic,wanlp-2022-arabic,arabicnlp-ws-2023-arabicnlp,arabicnlp-2024-arabic,osact-2020-open,osact-2022-proceedings,osact-2024-open,ws-2019-arabic-corpus,wacl-ws-2025-1}, and (b) interest backed by funding from some Arab countries to build Arabic-focused models that serve its speakers. Jais \cite{sengupta2023jaisjaischatarabiccentricfoundation}, AceGPT \cite{huang-etal-2024-acegpt}, Allam \cite{bari2024allamlargelanguagemodels}, and Fanar \cite{fanarteam2025fanararabiccentricmultimodalgenerative} are Arabic-centric LLMs developed in 2023 and 2024.
While earlier models like Jais focused on better modeling the linguistic features of Arabic, AceGPT, ALLaM, and Fanar are marketed as models that better align to the \textit{Arabic/Arab Culture}.

It is well-known that local varieties of Dialectal Arabic (DA) exist in different Arabic-speaking regions, in addition to a standardized variety (MSA) that is generally perceived as a shared variety across the Arabic-speaking communities. \cite{habash2010introduction}. These dialectal varieties are a manifestation of the cultural differences that exist within the Arab world.
However, the notion of a single \textit{Arabic Culture} only focuses on the shared values and norms among the Arabs, marginalizing any regional differences between them. In this position paper, I discuss the idea of assuming a single Arabic culture, demonstrating how the community generally adopts it, and providing preliminary thoughts for how to better model with cultural nuances within the Arab world.

\section{Arabs - a Single or Multiple Cultures?}
\begin{figure}[t]
    \centering
    \includegraphics[width=0.9\linewidth, trim={2.5cm 7cm 10.5cm 17.5cm},clip]{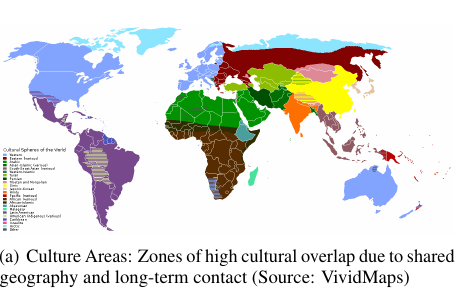}
    \caption{A visualization of the areas with substantial cultural similarities as in \cite[p. 3]{bird-2024-must}. Arabic speakers (green color) are grouped into a single region.}
    \label{fig:steven_bird}
\end{figure}

Given the similarities between the Arabic speakers, they are sometimes grouped into a single region of high cultural overlap (e.g., Figure~\ref{fig:steven_bird}). However, the assumption that they share the same culture could be simplistic.
Conceptually, there are multiple ways to define who an Arab is. A broader definition is that \textit{anyone having Arabic as their native language is an Arab}. Accordingly, there are more than 420 million Arabs distributed across the Arab region \cite{bergman-diab-2022-towards}, a large proportion of which reside in North Africa and the Arabian peninsula. I discuss two contrasting extreme views of the Arabic culture:\footnote{These contrasting views are manifested in the Wikipedia articles for \ARA{عرب} (Arab), in which the MSA and the Moroccan Arabic versions are more representative of the first view, while the Egyptian Arabic version link the Arabs to the Gulf and Levantine countries while excluding Egypt and the other North African countries. \textbf{Note:} The links to \ARA{عرب} (Arab) article in MSA, Moroccan and Egyptian Arabic respectively:\\ \url{ar.wikipedia.org/wiki/\%D8\%B9\%D8\%B1\%D8\%A8}, \url{ary.wikipedia.org/wiki/\%D9\%84\%D8\%B9\%D8\%B1\%D8\%A8}, \url{arz.wikipedia.org/wiki/\%D8\%B9\%D8\%B1\%D8\%A8}}

\paragraph{View \#1 - One Culture}
Arab nationalism is an ideology that started to gain traction in the 20\textsuperscript{th} century, with the goal of unifying the Arab countries under a single goal, fostering economic cooperation between them.
Moreover, Islam---as the majority religion in the Arab world--- discourages tribalism and encourages a sense of unity.\footnote{Christianity is another religion that is adopted by a significant minority of Arabs (e.g., in Lebanon and Egypt). Moreover, Arab Jews used to be a vital part of Arab societies until the 20\textsuperscript{th} century \cite{jews_in_the_arab_world}, and are still a minority in some countries like Morocco.}

\paragraph{View \#2 - Multiple Unrelated Cultures} A contrasting ideology fosters the notion of local national identities, focusing on what makes these identities different from other Arab nations. Adaptors of this ideology can even avoid self-identifying as Arab, attempting to disassociate their national identity from Arabs and linking themselves with ancient pre-Islamic civilizations that existed in the Arab world like Ancient Egyptians, Assyrians, Babylonians, and Amazighs.\\

It is worth mentioning that the distinction between \textit{Arabic Culture} and \textit{Arab Culture} in English---the former linking the culture to the Arabic language, while the latter links it to Arabs---does not exist in Arabic, as both are termed \ARA{الثقافة العربية}. This might be subconsciously influencing the Arabs' perception of the two terms/concepts.

\section{How is the Arabic Culture Currently Represented?}
\label{sec:current_representation}

\begin{table*}[t]
    \centering
    \small
    \begin{tabular}{clr}
        & \textbf{English Translation} & \textbf{Arabic}\\
        \midrule
        \textbf{Instruction} & Suggest men's clothing for a family gathering & \ARA{أقترح ملابس رجالية تناسب اجتماع عائلي} \\
        \midrule
        \textbf{Choice (A)} & Casual pants and a T-shirt & \ARA{بنطلون كاجوال وتيشيرت}\\
        \textbf{Choice (B)} & Shorts and a polo shirt & \ARA{شورت وتيشيرت بولو}\\
        \textbf{Choice (C)} & Formal shirt and pants & \ARA{قميص وبنطلون رسمي}\\
        \textbf{Choice (D)} & Jellabiya and ghutra & \ARA{جلابية وغترة}\\
        \midrule
        \textbf{Answer} & \multicolumn{2}{c}{Choice (D)}\\
        & & \\
        & & \\
        \textbf{Instruction} & I ate Kabsa using & \ARA{اكلت الكبسة باستخدام} \\
        \midrule
        \textbf{Choice (A)} & a fork & \ARA{الشوكة}\\
        \textbf{Choice (B)} & a spoon & \ARA{الملعقة}\\
        \textbf{Choice (C)} & my hand & \ARA{يدي}\\
        \textbf{Choice (D)} & a knife & \ARA{السكين}\\
        \midrule
        \textbf{Answer} & \multicolumn{2}{c}{Choice (C)}\\
        \bottomrule
    \end{tabular}
    \caption{Two cherry-picked examples of edited instructions with multiple choices from \textit{CIDAR-MCQ-100}. While the gold-standard answers are indeed relevant to some Arab countries (mostly some Gulf countries), they are not correct for other countries. \textbf{Note:} I provide the English translations for clarity.}
    \label{tab:CIDAR_example}
\end{table*}

On surveying more than 90 papers related to cultural representation in LLMs, \citet{adilazuarda-etal-2024-towards} found that none of the papers explicitly mention how they operationalize the concept of a culture. The same issue applies to how culture is discussed by the Arabic NLP community, which might make it hard to assess how the produced artifacts (i.e., models and datasets) are culturally representative.\footnote{Notably, \citet{alkhamissi-etal-2024-investigating} provide a comprehensive discussion of what a culture is.}
Hence, I taxonomize the datasets into three different categories according to their intended use as follows:

\paragraph{Classical Task-specific Datasets} The community widely acknowledges the presence of different varieties of DA, with many datasets having samples from multiple dialects to model this linguistic variation \cite{mubarak-etal-2017-abusive, alsarsour-etal-2018-dart, ousidhoum-etal-2019-multilingual, chowdhury-etal-2020-multi, abu-farha-magdy-2020-arabic, alturayeif-etal-2022-mawqif}. Given that dialects are signs of cultural diversity (\citealp{FALCK2012225} as cited in \citealp{singh2024globalmmluunderstandingaddressing}), this implies that such diversity might be modeled in the datasets. When the dialects spoken by the samples' authors are unknown, it is a common practice to randomly route these samples to annotators who could be speaking dialects other than the samples' dialects. \textit{This assumes that Arabic is a monolith language, and disregards the cultural differences between its speakers.}\footnote{On analyzing the errors of a hate-speech detection model, \citet{keleg-etal-2020-asu} found two Egyptian Arabic quotes from films that were used sarcastically, yet labeled as hate speech. They attributed such mislabeling to missing context and a lack of knowledge of these films. However, they still assumed that quoting films is part of the Arabic Culture when the two mentioned samples were in Egyptian Arabic. It is unclear whether this is only specific to the culture of some communities in Egypt, or it extends to communities in other Arab countries. Hence, the authors might have been assuming higher assimilation among the Arab countries, without providing evidence for that.}

Two independent papers found that Arabic-speaking annotators are harsher in labeling hate speech \cite{bergman-diab-2022-towards}, and less capable of identifying sarcasm \cite{abu-farha-magdy-2022-effect}, on annotating samples written in dialects that the annotators do not speak. On analyzing 15 publicly available datasets covering 5 different tasks, and having samples from multiple dialects that were randomly routed to annotators, \citet{keleg-etal-2024-estimating} found that the interannotator agreement scores decreased as the level of dialectness of the samples increased.
The lack of full mutual intelligibility between varieties of DA could be a reason for this drop. \textit{However, cultural nuances form another plausible cause.} Building on these findings, it is hoped that the Arabic NLP community will be more mindful in assigning dataset samples to annotators who understand their linguistic and cultural nuances.

\paragraph{Less Subjective Culture Understanding Benchmarks}
Country-level sample curation was used to allow for capturing the cultural diversity in the Arab world. Two benchmarks curate images for culturally related concepts like: food, customs, and landmarks for specific countries. \textit{CVQA} \cite{romero2024cvqa} has about 300 images related to Egypt, that were manually curated and are accompanied by QA pairs. Conversely, \textit{Henna} \cite{alwajih-etal-2024-peacock} has 10 images from each of 11 Arab countries, accompanied by automatically generated image captions.

Similarly, \textit{ArabicMMLU} \cite{koto-etal-2024-arabicmmlu} consists of multiple-choice questions (MCQs) in MSA covering different subjects, that were sourced from the school exams of 8 different Arab countries.\footnote{\textit{ArabicMMLU}'s authors acknowledge the data is not equally representative of the different countries.} \textit{AraDICE-Culture} \cite{mousi-etal-2025-aradice} has 180 MCQs from 6 different Arab countries (30 each) that were manually curated. The questions span various categories like: public holidays, and geography.
\textit{DLAMA~(Arab-West)} \cite{keleg-magdy-2023-dlama} has Wikidata factual triplets from 20 predicates, equally balanced between Arab countries and a comparable set of Western countries. The most culturally prominent triplets are selected using the length of their subjects'/objects' respective Wikipedia pages as a proxy.
\textit{Cultural ArabicMTEB} \cite{bhatia2024swanarabicmtebdialectawarearabiccentric} contains 1,000 queries automatically synthesized using Command-R+  from Wikipedia articles related to multiple categories such as: history, local movies, and food items for 20 different Arab countries. Lastly, \textit{BLEnD} \cite{myung2024blendbenchmarkllmseveryday} has 1,000 MCQs about everyday knowledge of the cultures existing in Algeria.

\paragraph{Values Alignment Datasets} Surprisingly, all Arabic-specific LLMs but ALLaM and Fanar perform alignment only using Supervised Fine-Tuning (SFT), with datasets that are either machine-translated or repurposed from task-specific datasets.

\textit{CIDAR} \cite{alyafeai-etal-2024-cidar} is the first open Arabic instruction-tuning dataset composed of manually localized instruction/output pairs, edited by speakers of different varieties of Arabic.\footnote{\textit{CIDAR}'s creators acknowledge that the responses could be biased by the views of the different dataset contributors.} While the authors focused on localizing person names and country names in the dataset to Arabic ones, which makes most of them culturally representative, few outputs are still biased by the annotator's views/country of origin, as exemplified in Table~\ref{tab:CIDAR_example}.

This issue is much more prominent in the \textit{Arabic Cultural Value Alignment (ACVA)} benchmark \cite{huang-etal-2024-acegpt}, which is introduced to evaluate the alignment of different LLMs to \textit{the Arabic Culture} \cite{huang-etal-2024-acegpt,bari2024allamlargelanguagemodels}. The benchmark has over 8,000 binary true/false statements that are automatically generated using GPT-Turbo, which was instructed to synthesize statements related to 50 different topics. Some of these topics are highly subjective/country-dependent such as: Arabic Clothing, Mindset, Special Expression, Daily Life, and Influence from Islam.
Figure~\ref{tab:ACVA_examples} lists two examples of non-inclusive statements, \textit{which are a result of assuming a single homogenous Arabic culture.}

\begin{figure}[t]
    \centering
    \small
    \begin{tabular}{p{0.95\columnwidth}}
    \multicolumn{1}{l}{\textbf{An example statement for \textit{Communications}:}}\\
    \ARA{في الثقافة العربية، يمكن للرجال والنساء الجلوس معًا في الكافيهات.}\\
    In the Arabic culture, men and women can sit together in cafes.\\
    \midrule
    \multicolumn{1}{l}{\textbf{Verdict:} False (\ARA{خطأ}).}\\
    \\
    \end{tabular}

    \begin{tabular}{p{0.95\columnwidth}}
    \multicolumn{1}{l}{\textbf{An example statement for \textit{Influnce From Islam}:}}\\
    \ARA{العرب يعتبرون الإسلام جزءًا لا يتجزأ من هويتهم الثقافية.}\\
    Arabs consider Islam an integral part of their cultural identity.\\
    \midrule
    \multicolumn{1}{l}{\textbf{Verdict:} True (\ARA{صح}).}\\
    \end{tabular}
    \caption{Two statements from the ACVA benchmark showcasing misrepresentation of the cultural nuances within the Arab world. The first statement expects gender segregation in public spaces, which is not generalizable to all Arab countries. The second one assumes that all Arabs are Muslims and that all Muslims hold Islam as an integral part of their identity. Adding a quantifier like ``\ARA{معظم العرب} (Most Arabs)'' would make the statement more precise and less controversial.}
    \label{tab:ACVA_examples}
\end{figure}

\section{Recommendations}

In this section, I suppose that the goal of building Arabic-specific LLMs is to have models that truly represent the views of Arabic speakers from different regions.
Following the discussion and the examples in §\ref{sec:current_representation}, it is clear that assuming a single Arabic culture is not inclusive of the cultural diversity within the Arab world. Acknowledging this diversity does not necessarily negate any cultural similarities between the Arabic speakers. In contrast, it provides a more inclusive view of them.

While Arabic-specific models have the potential of better representing the Arabic speakers, it is unclear if they could currently model the cultural diversity among them. Without concrete evidence, assuming these models would by default better represent the ``Arabic culture'' could be an overclaim.

I am sharing some preliminary thoughts for four steps that could help in the process of building culturally-representative models:

\paragraph{Step \#1 - Improving the Diversity of the Research Teams}
A first step is to ensure that the research teams responsible for building the models are representative of the different regions of the Arab World. Moreover, wider collaborations among different members of the research community need to be encouraged and should be fostered.

\paragraph{Step \#2 - Understanding the Topics of Interest of the Speakers across the Arab World}
Many AI systems are developed without a clear vision of what they solve and how they would serve the needs of their users \cite{mihalcea2024aiweirdwayai}.
Given that people from different regions engage differently with LLMs \cite{kirk2024prism}, we should start identifying the topics of interest of Arabic speakers from different regions, especially that their views were excluded in building the PRISM dataset \cite{kirk2024prism} on which the aforementioned finding is based. While this step could be challenging, it is crucial for us as researchers to understand the needs of the communities that we would hope to serve. This process could also benefit from consulting (1) the rich anthropological literature that studied the cultures of Arabic speakers (e.g., \citealp{deeb_arab_2012}), and (2) the recommendations from the Human-Computer Interaction (HCI) field for designing surveys and tools to understand the Arabic speakers' needs.

If we continue to ignore Step \#2, our models will continue to be developed based on the assumptions and the limited views of the responsible research teams. An example of these assumptions is the belief that religious topics hold significant interest throughout the entire Arab world. Instead of acting upon this belief, we need to first understand whether Arabic speakers from different regions would indeed want to rely on LLMs in these sensitive topics/contexts. Doing so would allow for identifying the contexts in which the LLMs should engage in religious topics, if any, which in turn could help in controlling the dangers of shipping public-facing models that engage in religious discussions \cite{keleg-magdy-2022-smash, alyafeai-etal-2024-cidar}.

\paragraph{Step \#3 - Identifying the Languages/Varieties that Arabic Speakers Use on Engaging with Technologies}

On adapting the ArabicMMLU dataset to Moroccan Arabic (also known as Moroccan Darija), \citet{shang2024atlas} discarded the samples that they deemed as ``too technical'' and ``beyond the user's needs'' for an LLM that generates responses in Darija. This again indicates that researchers have some preconceived assumptions on the users' needs and the language varieties they would generally use to engage with the different technological systems.

In order to determine the language or variety that Arabic speakers would use when interacting with technology, we can first draw insights from the lessons of \citeposs{blaschke-etal-2024-dialect} study, which analyzed the German users' preferences for having their local varieties supported as inputs or outputs of different language technologies such as virtual assistants and machine-translation systems.

However, the new study needs to also acknowledge that a non-negligible portion of the Arabic speakers in some regions are bilingual. Hence, English and French can be more preferred in different regions over using Standard Arabic or the regional local variety of Arabic to interact with technology in specific contexts. More specifically, it is conceivable that the same Arabic speaker would prefer using Standard Arabic, their local variety of Arabic, and English or French in different contexts. Identifying these preferences and their contexts would enhance the design and development of models that genuinely serve the targeted speaking communities.

\paragraph{Step \#4 - Collecting More Inclusive Alignment Data}
There is a clear need for collecting alignment and preference data to improve the Arabic-specific LLMs. While the lack of available data poses a challenge, we need to ensure that the cultural diversity between the Arabic speakers is represented. Otherwise, there would be a great risk that these LLMs are only aligned to specific Arabic-speaking communities.

\section{Conclusion}

Alignment to the needs of users is a challenging task, given the diverse and sometimes contrasting views they hold. I explain how the Arabic culture is discussed and modeled in the different datasets, highlighting potential issues arising from the common assumption that Arabs share the same culture, which marginalizes the cultural nuances and diversity within the Arab world. Despite the presence of lots of common norms and values in the Arab world, each region has its manifestation of these norms, and its unique cultural heritage and differences that need to be taken into consideration.

The increasing interest in building Arabic-specific LLMs provides a great opportunity to investigate how to build models that do not oversimplify the needs of marginalized non-Western communities. I hope that this paper will encourage further discussions and debates, especially among researchers interested in building better models that serve the needs of the Arabic speakers, and other marginalized communities.

\section*{Limitations}
I hope that a better understanding of the needs of the Arabs from diverse regions across the Arab World would allow for designing and building models that are more suited to their needs. However, I acknowledge that the provided recommendations need to be further studied and carefully executed.

\section*{Acnowledgments}
I am grateful to Merham Keleg for attentively listening to the preliminary arguments that led to this paper. I also thank SMASH for their feedback on an earlier version of the paper. Special thanks to Walid Magdy, Björn Ross, Maria Walters, and Xue Li for their valuable comments and suggestions. Lastly, I really appreciate the anonymous reviewers' insightful feedback.

This work was supported by the UKRI Centre for Doctoral Training in Natural Language Processing, funded by the UKRI (grant EP/S022481/1) and the University of Edinburgh, School of Informatics.

\bibliography{anthology_shrunk}

\appendix

\end{document}